\documentclass{article}

\usepackage{amsmath,graphicx}
\usepackage{spconf}
\usepackage{cite}
\usepackage{array}
\usepackage{multirow}
\usepackage{balance}
\usepackage{epstopdf}

\newcolumntype{x}[1]{%
>{\centering\hspace{0pt}}p{#1}}%

\title{DefectNET: multi-class fault detection on highly-imbalanced datasets}
\name{N. Anantrasirichai and David Bull\thanks{This work was supported by Innovate UK: Palantir - Real time inspection and assessment of wind turbine blade health.}}
\address{Visual Information Laboratory, University of Bristol, UK}
\begin{document}
%
\maketitle
\begin{abstract}
As a data-driven method, the performance of deep convolutional neural networks (CNN) relies heavily on training data. The prediction results of traditional networks give a bias toward larger classes, which tend to be the background in the semantic segmentation task. This becomes a major problem for fault detection, where the targets appear very small on the images and vary in both types and sizes. In this paper we propose a new network architecture, DefectNet, that offers multi-class (including but not limited to) defect detection on highly-imbalanced datasets. DefectNet consists of two parallel paths, which are a fully convolutional network and a dilated convolutional network to detect large and small objects respectively. We  propose a hybrid loss maximising the usefulness of a dice loss and a cross entropy loss, and we also employ the leaky rectified linear unit (ReLU) to deal with rare occurrence of some targets in training batches. The prediction results show that our DefectNet outperforms state-of-the-art networks for detecting multi-class defects with the average accuracy improvement of  approximately 10\% on a wind turbine.

\end{abstract}
\section{Introduction}
\label{sec:intro}

\begin{figure}[t!]
	\centering
 		\includegraphics[width=\columnwidth]{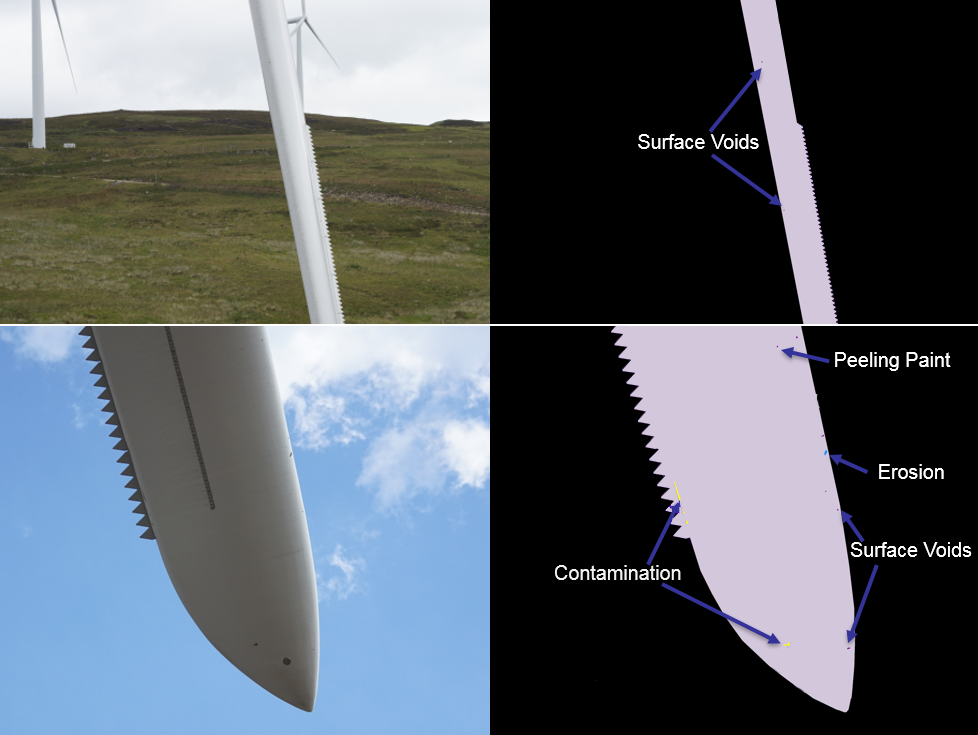}
	\caption{\small Examples of a highly-varying dataset of defects on wind turbine blade images (left) with ground truth (right). The top image was captured from a distance, causing detects to appear to be significantly smaller than those in the bottom image. }
	\label{fig:examplebladesmask}
\end{figure}

Vision-based inspection for anomalous objects or faults has become an attractive tool as it offers cost effectiveness, potentially real-time operation and automation that deals with continuously generated data.  In general, detection of such anomalies involves two tasks, namely segmentation and classification. Machine learning techniques, particularly recent popular deep learning methods, have been proposed to achieve both processes simultaneously, referred to as semantic segmentation \cite{Shelhamer:FCN:2017}.  However, their performance of detecting multiple faults on very large background is still limited due to the variations of fault types, their sizes and uncontrolled lighting for outdoor image acquisition. Moreover, the frequency of occurrence of each fault type varies, leading to highly imbalanced datasets. This causes classifier algorithms to bias classes which have a greater number of instances and preferentially predict majority class data. In this paper, we use the important application of detecting defects on wind turbines (Fig. \ref{fig:examplebladesmask}) as a vehicle for addressing this problem.

Numerous approaches have been proposed to create balanced distributions of data using data manipulation \cite{He:Learning:2009}. These include downsampling majority classes, oversampling minority classes, or both. However, for semantic segmentation, augmenting the number of samples in the smaller classes always includes more of the surrounding areas, resulting in more samples of the larger class too -- augmenting the small classes always add more background samples. Hence we employ the other techniques that deal with the imbalanced training data issue by modifying the network architecture and the validation loss.

For convolutional neural networks (CNNs), classical techniques balance classes using weights computed by analysing dataset statistics. A dice coefficient-based loss has been employed by Milletari et al. \cite{Milletari:vnet:2016}, giving better segmentation results than re-weighted soft-max with loss.  Later, the generalised dice loss was proposed for multiple classes by Sudre et al. \cite{Sudre:generalised:2017}. Although the dice loss has become popular for medical imaging tasks, the cross entropy loss has been reported to have better performance in some applications \cite{zpolina:Comparison:2018}. Recently the ideas of combined loss have been introduced \cite{Graham:XY:2018, Isensee:nnU-Net:2018}. However none of these approaches have been tested in a challenging scenario, such as defect detection, where each training batch does not contain all class samples.

Another problem when detecting small objects using CNNs is the decrease of receptive field size as the network depth increases. CNNs used for  classification employ a pooling layer after the convolutional layer to combine the outputs of neuron clusters from one layer into a single neuron in the next layer. This results in feature maps with reduced spatial resolution with the outcome that very small areas could be represented by only one pixel. Dilated convolutional layers have therefore been introduced to enlarge the field of view of the filters \cite{Chen:DeepLab:2018}. Generally the dilation rate of the filter size (inserting more zeros) increases with larger depth. However this can cause false positives for small classes \cite{Li:CSRNet:2018}. We therefore diminish this problem here by combining feature maps from traditional convolutional layers with dilated convolutional layers.

In this paper, we propose a new network, DefectNet, for detecting various defect size on a highly imbalanced dataset. 
The proposed DefectNet, shown in Fig. \ref{fig:DefectNet}, has two parallel paths, where one of them makes use of skip layers \cite{Shelhamer:FCN:2017} to detect medium to large objects. The additional path employs the dilated convolutional layers to increase receptive field size for small object defect detection. The two paths are combined with sum fusion at the construction end.
We also propose a hybrid loss function that combines the advantages of the cross entropy and dice losses. The cross entropy loss compensates for some absent classes in the training batches, whilst the dice loss improves the balance between the precision and recall. We employ LeakyReLU (Leaky rectified linear unit) to prevent zero gradients when some small classes are not included in several training batches consecutively. 

Although the example application is defect detection on wind turbine images, our approach should be useful in general as various applications across a wide range of fields experience problems associated with highly imbalanced classes, e.g. segmentation on medical images \cite{Ronneberger:Unet:2015},   radio frequency interference mitigation \cite{Akeret:Radio:2017}, event classification of eye movement \cite{Anantrasirichai:fixation:2016}, hazard monitoring \cite{Anantrasirichai:Application:2018}. 

The remainder of this paper is organised as follows.  The proposed DefectNet is described in Section \ref{sec:DefectNet}. The performance of the method is evaluated in Section \ref{sec:results}. Finally, Section \ref{sec:conclusion} presents the conclusions of this work.

\section{DefectNet}
\label{sec:DefectNet}

\begin{figure}[t!]
	\centering
 		\includegraphics[width=\columnwidth]{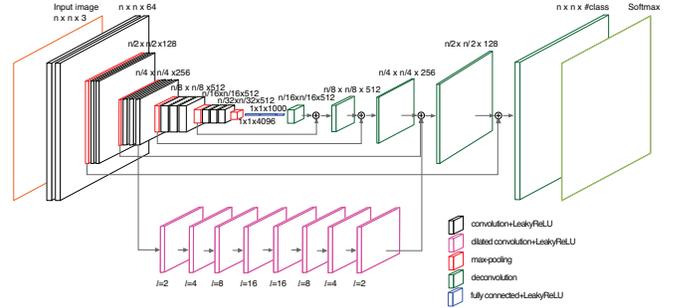}
	\caption{\small DefectNet based on the VGG-19 architectures with additional path of dilated convolution with dilation factors of 2, 4, 8, 16, 16, 8, 4, 2 from low to high levels repectively.}
	\label{fig:DefectNet}
\end{figure}

The architecture of  DefectNet is illustrated in Fig. \ref{fig:DefectNet}. It combines two paths able to detect different target sizes. The first path makes use of the VGG-19 architecture and skip layer fusion creating fully convolutional network. It is an enhanced version of the architecture of the FCN-8 network \cite{Shelhamer:FCN:2017}, introducing additional skip layers in pool1 and pool2 in order to include the low-level features for finer prediction. The second path employs the dilated convolution (further described in Section \ref{ssec:dilated}) to detect the small objects.
All filters of the convolutional layers have a kernel size of 3$\times$3, but those of the second path have zero-inserts to create dilated convolution. All convolution processes operate with the stride of 1. The following subsections highlight the differences of the proposed deep learning network to the traditional ones.

\subsection{Dilated convolution}
\label{ssec:dilated}

Dilated convolution, also referred to as atrous convolution, enlarges the field of view of the filters to incorporate larger context by  expanding the receptive field without loss of resolution.
The dilated convolutions are defined  in Eq.\ref{eqn:dilatedconv} \cite{Yu:Multi:2016}, where $F$ is a feature map, $k$ is a filter, $\ast_l$ is a convolution operator with a dilation factor $l$, meaning ($l-$1) zeros are inserted. We construct eight dilated convolution layers (coloured pink in Fig. \ref{fig:DefectNet}) after the second group of convolutional layers of the other FCN path. The dilation factor $l$ is defined in Fig. \ref{fig:DefectNet}. This structure is similar to the basic context network architecture proposed in \cite{Yu:Multi:2016}, but we add more dilated convolution layers with decreasing $l$. This improves local low-level features, where spatial relationships amongst adjacent pixels may be ignored due to sparsity of the dilated filters when increasing $l$ \cite{Hamaguchi:Effective:2017}.
\begin{equation}
\label{eqn:dilatedconv}
	 (F\ast_l k)(\textbf{p}) = \sum_{\textbf{s} + l \textbf{t} = \textbf{p}} F(\textbf{s}) k(\textbf{t})
\end{equation}

\subsection{Hybrid loss}
\label{ssec:hybridloss}

The most commonly used loss function for the task of image segmentation is a pixel-wise cross entropy loss. The class weighting technique is employed to balance the classes to improve training. Another popular loss is the dice loss. The dice coefficient performs better on class imbalanced problems because it maximises a metric that directly measures the region intersection over union. The gradients of the dice coefficients are however not as smooth as those of cross entropy, particularly when both predictions and labels are close to zero. For our dataset, the data in each batch do not contain all classes, leading to very noisy  training error. The cross entropy loss, on the other hand, allows the absent classes to affect the backpropagation less than the dice loss. Therefore we propose the hybrid loss defined in Eq. \ref{eqn:hybridLoss}, where the weighted cross entropy loss $L_{wce}$ and generalised dice loss $L_{gdice}$ are merged in the proportion of the number of class samples present in the training batch (Eq. \ref{eqn:weightgamma}).
\begin{equation}
\label{eqn:hybridLoss}
	 L_{hybrid} = (1- \gamma) \max(1,L_{wce}) + \gamma L_{gdice}
\end{equation}
\begin{equation}
\label{eqn:weightgamma}
	 \gamma =  \frac{1}{K} \sum^K_{c=1} \left( ( \sum^M_{m=1}  t_{cm}) > 0 \right)
\end{equation}

The loss $L_{wce}$ is definded as $L_{wce}=-\sum^K_{c=1} w_c  t_c log(y_c) $, where $K$ is the number of classes and  $y_{c}$ and $t_{c}$ represent the prediction and target, respectively. Generally the weights are computed from pixel counts, i.e. a weight of class $c$ is $w_c = 1 / f_c$, where $f_c$ is pixel counts of class $c$ \cite{Long:fully:2015}. However applications such as defect detection have highly-imbalanced classes. Their scores could approach zero and cause poor performance because significantly higher weight values are given to the smaller classes. We hence propose to use $w_c = 1 / \sqrt{f_c}$.
Equation \ref{eqn:gdiceloss} shows the generalized dice loss proposed in \cite{Sudre:generalised:2017}, where $y_{cm}$ and $t_{cm}$ represent the prediction and groundtruth of the pixel $m$ from the total $M$ pixels. The weight in the generalized dice loss is generally calculated as the inverse of the squared sum of all pixels of class $c$, but this will give priority to the small and absent classes. Therefore in DefectNet, we adapt the dice loss with the same $w_c$ used in $L_{wce}$.
\begin{equation}
\label{eqn:gdiceloss}
	 L_{gdice} = 1- \frac{2 \sum^K_{c=1} w_c \sum^M_{m=1} (y_{cm} t_{cm})}{\sum^K_{c=1} w_c \sum^M_{m=1} (y^2_{cm} + t^2_{cm})}
\end{equation}

\subsection{Leaky rectified linear unit}
\label{ssec:classweighting}

Because an entire dataset cannot be fed into the neural network at once, it is divided into multiple batches with random data selection. This means some classes may not present in particular batches, causing some nodes within the layers to have an activation of zero. The backpropogation process does not update the weights for these nodes because their gradient is also zero. This is called the `dying ReLU' problem. To prevent this, we employ the leaky ReLU activation function. It allows a small gradient when the unit is not active (negative inputs) so that the backpropogation will always update the weights. The leaky ReLU is defined as shown in Eq. \ref{eqn:leakyrelu} \cite{He:Delving:2015} and $\alpha$ is set to 0.1 during training.
\begin{equation}
\label{eqn:leakyrelu}
	 f(x) = \begin{cases}
  x & \text{if $x>0$} \\
  \alpha x & \text{otherwise}
\end{cases}
\end{equation}

\begin{table*}
\caption{Classification performances of CNN models on defect detection. The average accuracies are computed from 6 defects and previously repaired masks. The accuracies are shown in percentage and the processing time are shown in ms/patch. }
 \centering
\scriptsize
 \begin{tabular}{lccccccccccc }
 \hline
\multirow{ 2}{*}{Network} & \multicolumn{9}{c}{Class accuracy}& Avg acc.  & Time \\ \cline{2-10}
& BG & Blade & Erosion & Repairs & Contamination & Peeling paint & Scratches & Surface voids & Chipped paint & of defects & (ms)\\
\hline
FCN ($L_{wce}$)& 99.92 & 99.97 & 80.48 & 20.30 & 52.48 & 17.34 & 21.09 & 33.53 & 23.50 & 35.53 & 52.35 \\
FCN dilated ($L_{wce}$)& 99.90 & 99.87 & \textbf{97.29} & 59.54 & 55.29 & 52.51 & 82.57 & 69.62 & 33.93 & 64.39 & 58.48 \\
UNET ($L_{gdice}$) & 99.92 & \textbf{99.98} & 92.14 & 78.34 & 56.41 & 43.25 & 56.20 & 52.86 & 42.81 & 60.28 & 44.27 \\
deepLabv3+ ($L_{wce}$)& 99.95 & 99.95 &95.10 & 88.45 & 76.85 & 51.04 & 74.15 & \textbf{73.52} & 53.44 & 73.22 & 60.42 \\
\hline
DefectNet ($L_{ce}$)  & \textbf{99.97} & 99.92 & 74.18 & 14.30 & 32.36 & 14.15 & 9.79 & 25.16 & 33.04 & 29.00 & - \\
DefectNet ($L_{wce}$)    & 99.82 & 99.85 &  96.39 & 66.84 & 84.55 & 65.56 & 80.86 & 70.71 & 48.09 & 73.29 & - \\
DefectNet ($L_{gdice}$)   & 99.81& 99.81 & 89.01 & 57.19 & 62.21 & \textbf{77.28} & 69.47 & 68.33 & 54.83 & 68.33 & -\\ 
DefectNet ($L_{hybrid}$) & 99.82 & 99.80 & 95.62 & \textbf{90.10} & \textbf{84.75} & 74.57 & \textbf{84.04} & 70.94 & \textbf{57.97} & \textbf{79.71} & 62.83\\
\hline
 \end{tabular}
\label{tab:Classification}
\end{table*}

\begin{figure}[t!]
	\centering
 		\includegraphics[width=\columnwidth]{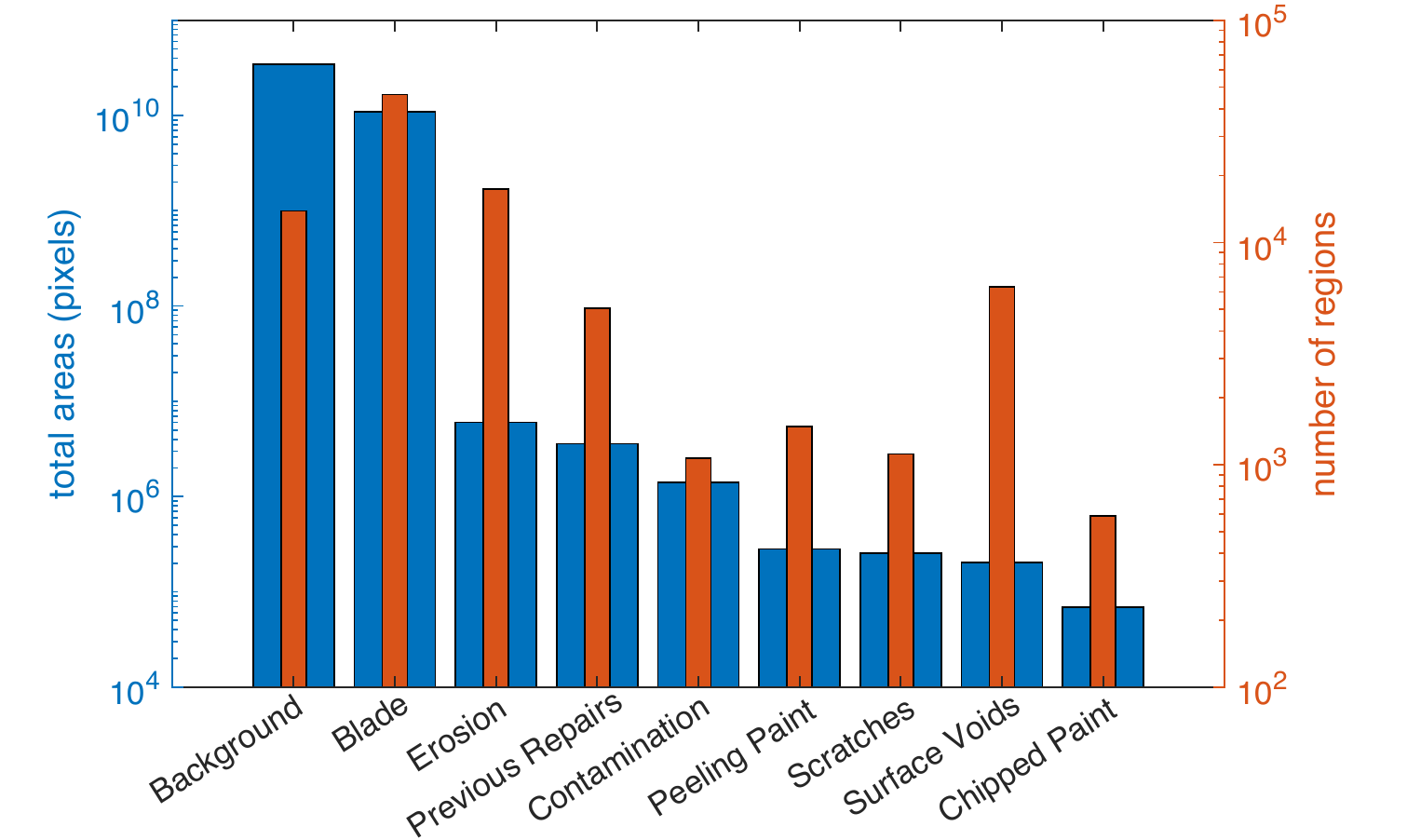}
	\caption{\small The areas in pixels and the number of regions of 9 classes we want to detect. }
	\label{fig:totalAreasVSNumRegions}
\end{figure} 
\begin{figure}[t!]
	\centering
 		\includegraphics[width=\columnwidth]{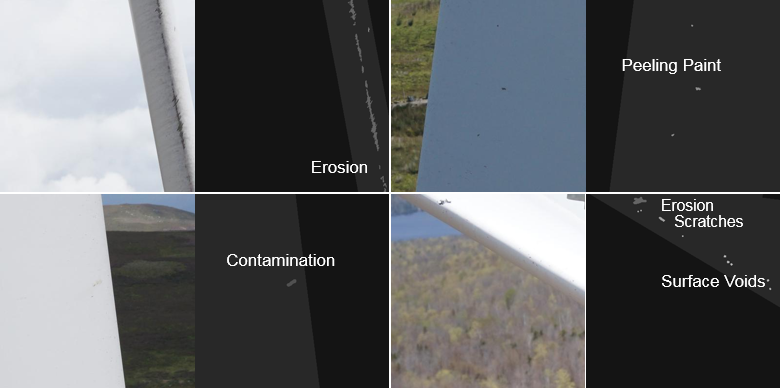}
	\caption{\small Examples of 400$\times$400 training patches. The left and the right of each pair are colour patches and ground truth labels.}
	\label{fig:trainingpatches}
\end{figure} 

\section{Results and discussion}
\label{sec:results}

The defect dataset contains a total of 2,188 images. Two thirds of the images were used for training and the remainder for testing. Nine classes were segmented and identified. The statistics of their areas and sizes are shown in Fig. \ref{fig:totalAreasVSNumRegions}. The blue and red bars show the number of pixels and regions of each class, respectively. The majority of segments were classed as background (BG) and blades. The most common defect type is erosion. Surface voids are the smallest targets, whilst chipped paint is the rarest class. The backgrounds include various contents, e.g. sky, sea, vegetation and other objects, such as distant wind turbines and houses. 

The original images vary in size and the blades can be very close or far from the cameras. In this experiment, each image was downsampled to 2000$\times$3000 pixels for computational purposes. For training, each image was divided into overlapping 400$\times$400 patches (Fig. \ref{fig:trainingpatches}) with each patch shifted vertically and/or horizontally by 20 pixels. If a patch only contains 2 classes, it will not be used. 
In total we have 61,250 training patches. We initialised weights and biases with the VGG-19 network trained on the ImageNet database \cite{Russakovsky:ImageNet:2015}. For testing, each image is also divided into patches that overlap by 200 pixels. The probability output maps are merged using an equally weighted average and the class with highest probability is labelled.

The results were compared with three state-of-the-art of semantic segmentation approaches, namely FCN \cite{Shelhamer:FCN:2017}, UNET \cite{Ronneberger:Unet:2015} and deepLabv3+ (Encoder-Decoder with dilated convolution) by Google \cite{Chen:Encoder:2018}. We applied weights $w_c = 1 / \sqrt{f_c}$ to the loss functions of FCN and deepLabv3+, because it achieved better results than using $w_c = 1$ or $w_c = 1 /{f_c}$. We employed the dice loss for UNET as it gave the best result, but the leaky ReLU was used because the dying ReLU created big variations in dice scores.  We also show the results of the dilated convolution path alone (FCN dilated) with our proposed loss function. When any network was trained using the cross entropy loss without weights $L_{ce}$, the performances was very poor.  All networks were trained on BlueCrystal (High Performance Computing facility with NVIDIA Tesla P100) at the University of Bristol. A batch size was set to 10 which was limited by system memory. The Adam optimiser was employed with an initial learning rate of 10$^{-4}$. We evaluated several dropout values for preventing the potential overfitting problem but observed no significant differences in performance.

\begin{figure}[t!]
	\centering
 		\includegraphics[width=\columnwidth]{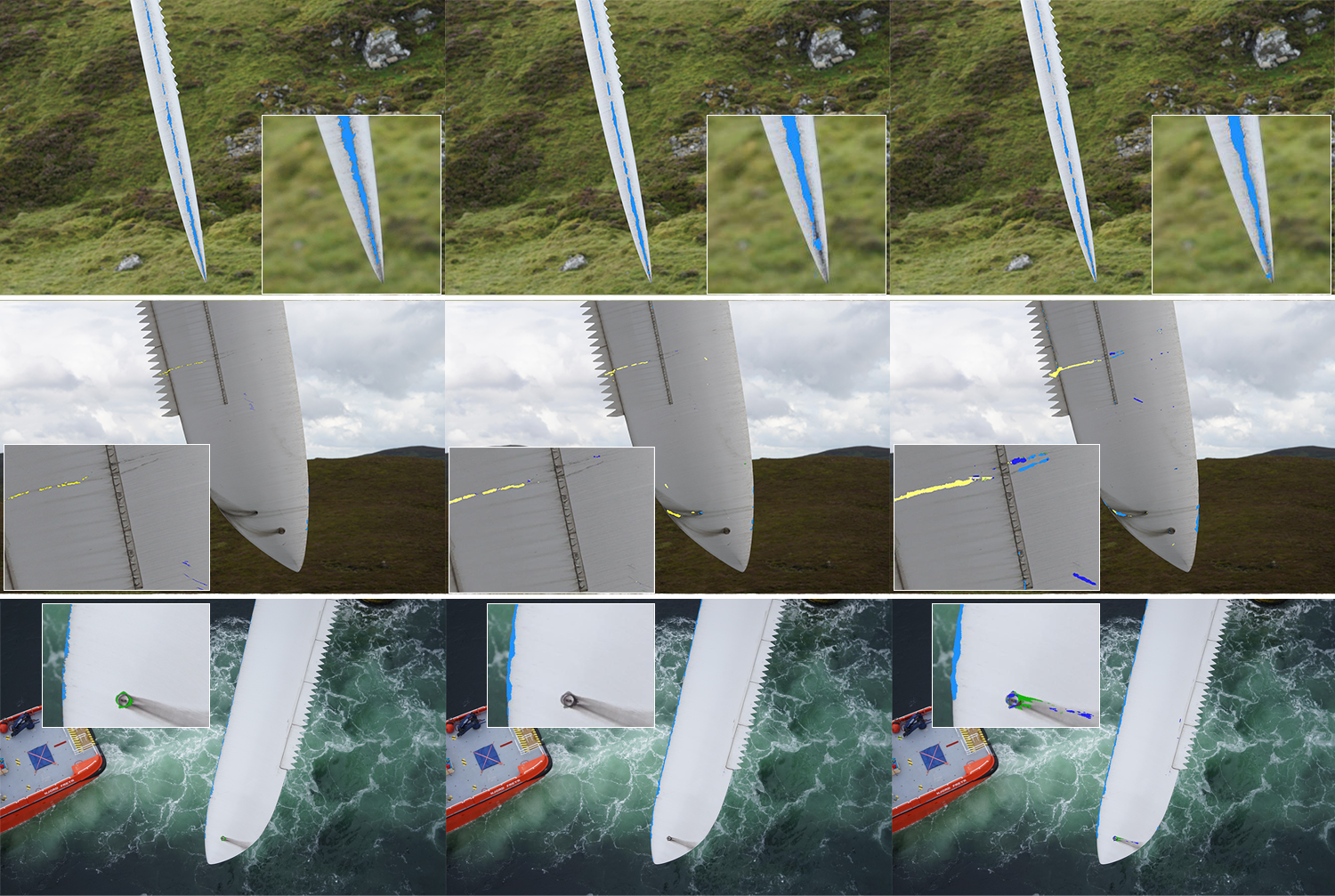}
	\caption{\small Predicted results by deepLabv3+ and DefectNet are shown in the middle and right columns respectively. The ground truths are shown in the left column. Erosion, scratch, contamination, surface void are masked with light blue, green, dark blue and yellow, respectively.   }
	\label{fig:results_false}
\end{figure}

Table \ref{tab:Classification} shows the accuracy of each detected category. The average accuracy is computed from 6 defects along with the previous repairs (as these are more important for monitoring the blade health of the wind turbines). It can be observed that DefectNet+$L_{hybrid}$ outperforms other networks in overall. However, it requires the most execution time in the prediction process ($\sim$4\% more than deepLabv3+). DefectNet+$L_{hybrid}$ may not be the best detector for some defects, but is the most consistent and gives the accuracies within 2.8\% of the most accurate techniques.
The dilated convolution can detect small objects better than traditional convolution with pooling (the accuracy of FCN dilated is almost twice that of FCN).
 deepLabv3+ and DefectNet+$L_{wce}$  were trained with the same loss function and their average accuracies were very close. However, the variance amongst class accuracies of DefectNet+$L_{wce}$ is lower ($\sim$9\%), i.e. the network is less biased to specific classes. We also show that the proposed hybrid loss can improve the detection performance by 9\% and 16\% over  $L_{wce}$ and $L_{gdice}$, respectively. 
Fig. \ref{fig:results_false} shows three subjective results. The magnified regions show that DefectNet incorrectly identified pixels of blades as defects more than deepLabv3+. However it missed defects less than deepLabv3+, which is more important for inspection of wind turbine blade health (false negatives is more serious than false positives in most applications of fault detection).

\section{Conclusion}
\label{sec:conclusion}

This paper presents a new architecture, DefectNet, for multi-class fault detection, where the numbers of class samples vary significantly.  We address the problem of a highly-imbalanced dataset using two parallel CNN paths to detect different target sizes. Large targets are detected using a modified FCN based on the VGG-19 structure and skip layers. Small targets are detected using dilated convolutional layers replacing the pooling layers that reduce the resolution of feature maps. To deal with the problem when some classes are absent in several successive training batches, a leaky ReLU is employed. We also proposed a hybrid loss that combines the weighted cross entropy and the generalised dice loss adaptively. The proposed DefectNet outperforms the state-of-the-art networks when detecting the defects on the blades of wind turbines by 9$-$32\%.

\vfill\pagebreak

\balance
\bibliographystyle{IEEEbib}
\bibliography{literature_review}

\begin{thebibliography}{10}

\bibitem{Shelhamer:FCN:2017}
E.~Shelhamer, J.~Long, and T.~Darrell,
\newblock ``Fully convolutional networks for semantic segmentation,''
\newblock {\em IEEE Transactions on Pattern Analysis and Machine Intelligence},
  vol. 39, no. 4, pp. 640--651, April 2017.

\bibitem{He:Learning:2009}
H.~He and E.~A. Garcia,
\newblock ``Learning from imbalanced data,''
\newblock {\em IEEE Transactions on Knowledge and Data Engineering}, vol. 21,
  no. 9, pp. 1263--1284, Sept 2009.

\bibitem{Milletari:vnet:2016}
F.~Milletari, N.~Navab, and S.~Ahmadi,
\newblock ``V-net: Fully convolutional neural networks for volumetric medical
  image segmentation,''
\newblock in {\em 2016 Fourth International Conference on 3D Vision (3DV)}, Oct
  2016, pp. 565--571.

\bibitem{Sudre:generalised:2017}
C.~H. Sudre, W.~Li, T.~Vercauteren, S.~Ourselin, and M.~Cardoso,
\newblock ``Generalised dice overlap as a deep learning loss function for
  highly unbalanced segmentations,''
\newblock in {\em Deep Learning in Medical Image Analysis and Multimodal
  Learning for Clinical Decision Support}. 2017, pp. 240--248, Springer
  International Publishing.

\bibitem{zpolina:Comparison:2018}
P.~Zablotskaia,
\newblock ``Comparison of cross-entropy and dice loss for cell nuclei detection
  via deep convolutional neural networks,'' 2018.

\bibitem{Graham:XY:2018}
S.~Graham, Q.~D. Vu, S.~E.~A. Raza, J.~T. Kwak, and N.~Rajpoot,
\newblock ``{XY} network for nuclear segmentation in multi-tissue histology
  images,''
\newblock {\em arXiv preprint arXiv:1812.06499}, 2018.

\bibitem{Isensee:nnU-Net:2018}
F.~Isensee, J.~Petersen, A.~Klein, D.~Zimmerer, P.~F. Jaeger, S.~Kohl,
  J.~Wasserthal, G.~Koehler, T.~Norajitra, S.~Wirkert, and K.~H. Maier-Hein,
\newblock ``nn{U}-{N}et: Self-adapting framework for u-net-based medical image
  segmentation,''
\newblock {\em arXiv preprint arXiv:1809.10486}, 2018.

\bibitem{Chen:DeepLab:2018}
L.~Chen, G.~Papandreou, I.~Kokkinos, K.~Murphy, and A.~L. Yuille,
\newblock ``Deeplab: Semantic image segmentation with deep convolutional nets,
  atrous convolution, and fully connected crfs,''
\newblock {\em IEEE Transactions on Pattern Analysis and Machine Intelligence},
  vol. 40, no. 4, pp. 834--848, April 2018.

\bibitem{Li:CSRNet:2018}
Y.~Li, X.~Zhang, and D.~Chen,
\newblock ``Csrnet: Dilated convolutional neural networks for understanding the
  highly congested scenes,''
\newblock in {\em 2018 IEEE/CVF Conference on Computer Vision and Pattern
  Recognition}, June 2018, pp. 1091--1100.

\bibitem{Ronneberger:Unet:2015}
O.~Ronneberger, P.~Fischer, and T.~Brox,
\newblock ``U-net: Convolutional networks for biomedical image segmentation,''
\newblock in {\em International Conference on Medical Image Computing and
  Computer-Assisted Intervention}. 2015, pp. 234--241, Springer.

\bibitem{Akeret:Radio:2017}
J.~Akeret, C.~Chang, A.~Lucchi, and A.~Refregier,
\newblock ``Radio frequency interference mitigation using deep convolutional
  neural networks,''
\newblock {\em Astronomy and Computing}, vol. 18, pp. 35--39, 2017.

\bibitem{Anantrasirichai:fixation:2016}
N.~Anantrasirichai, I.~D. Gilchrist, and D.~R. Bull,
\newblock ``Fixation identification for low-sample-rate mobile eye trackers,''
\newblock in {\em IEEE International Conference on Image Processing (ICIP)},
  Sept 2016, pp. 3126--3130.

\bibitem{Anantrasirichai:Application:2018}
N.~Anantrasirichai, J.~Biggs, F.~Albino, P.~Hill, and D.~Bull,
\newblock ``Application of machine learning to classification of volcanic
  deformation in routinely-generated insar data,''
\newblock {\em Journal of Geophysical Research: Solid Earth}, vol. 123, pp.
  1--15, 2018.

\bibitem{Yu:Multi:2016}
F.~Yu and V.~Koltun,
\newblock ``Multi-scale context aggregation by dilated convolutions,''
\newblock in {\em International Conference on Learning Representations}, 2016.

\bibitem{Hamaguchi:Effective:2017}
R.~Hamaguchi, A.~Fujita, K.~Nemoto, T.~Imaizumi, and S.~Hikosaka,
\newblock ``Effective use of dilated convolutions for segmenting small object
  instances in remote sensing imagery,''
\newblock {\em arXiv:1709.00179}, vol. 22, pp. 1--10, 2017.

\bibitem{Long:fully:2015}
J.~Long, E.~Shelhamer, and T.~Darrell,
\newblock ``Fully convolutional networks for semantic segmentation,''
\newblock in {\em IEEE Conference on Computer Vision and Pattern Recognition
  (CVPR)}, June 2015, pp. 3431--3440.

\bibitem{He:Delving:2015}
K.~He, X.~Zhang, S.~Ren, and J.~Sun,
\newblock ``Delving deep into rectifiers: Surpassing human-level performance on
  imagenet classification,''
\newblock in {\em IEEE International Conference on Computer Vision (ICCV)}, Dec
  2015, pp. 1026--1034.

\bibitem{Russakovsky:ImageNet:2015}
O.~Russakovsky, J.~Deng, H.~Su, J.~Krause, S.~Satheesh, S.~Ma, Z.~Huang,
  A.~Karpathy, A.~Khosla, M.~Bernstein, A.~C. Berg, and L.~Fei-Fei,
\newblock ``{ImageNet Large Scale Visual Recognition Challenge},''
\newblock {\em International Journal of Computer Vision (IJCV)}, vol. 115, no.
  3, pp. 211--252, 2015.

\bibitem{Chen:Encoder:2018}
L.-C. Chen, Y.~Zhu, G.~Papandreou, F.~Schroff, and H.~Adam,
\newblock ``Encoder-decoder with atrous separable convolution for semantic
  image segmentation,''
\newblock in {\em ECCV}, 2018.

\end{thebibliography}

\end{document}